\documentclass[letterpaper, 10 pt, conference]{ieeeconf}  
\IEEEoverridecommandlockouts                              

\usepackage{times} 
\usepackage{multicol}
\usepackage[bookmarks=true]{hyperref}
\usepackage{amssymb,amsmath}
\usepackage{graphicx}        
\usepackage[boxed]{algorithm2e}
\usepackage{subfigure}

\DeclareGraphicsExtensions{.pdf,.jpg,.png}

\newcommand{\Prob} {\ensuremath \mathbf{P}  }
\newcommand{\Exp} {\ensuremath \mathbb{E}  }

\graphicspath{{figs/}}

\title{\LARGE \bf
Gibbs Sampling Strategies for Semantic Perception of Streaming Video Data 
}

\author{Yogesh Girdhar$^{1}$ and Gregory Dudek$^{2}$
\thanks{$^{1}$Yogesh Girdhar is at Deep Submergence Laboratory, Woods Hole Oceanographic Institution, Woods Hole, MA 02542, USA 
        {\tt\small  yogi@whoi.edu}}%
\thanks{$^{2}$Gregory Dudek is at Center for Intelligent Machines, McGill University, Montreal, QC H3A0E9, Canada
        {\tt\small  dudek@cim.mcgill.ca}}%
}

\begin{document}

\maketitle
\thispagestyle{empty}
\pagestyle{empty}

\begin{abstract}
Topic modeling of streaming sensor data can be used for high level perception of the environment by a mobile robot. In this paper we compare various Gibbs sampling strategies for topic modeling of streaming spatiotemporal data, such as video captured by a mobile robot. Compared to previous work on online topic modeling, such as o-LDA and incremental LDA, we show that the proposed technique results in lower online and final perplexity, given the realtime constraints.
\end{abstract}

\section{Introduction}
Making decisions based on the environmental context of a robot's locations requires that we first model the context of the robot observations, which in turn might correspond to various semantic or conceptually higher level entities that compose the world. If we are given an observation model of these entities that compose the world then it is easy to describe a given scene in terms of these entities using this model; likewise, if we are given a labeling of the world in terms of these entities, then it is easy to compute the observation model for each individual entity. The challenge comes from doing these two tasks together, unsupervised, and with no prior information. ROST~\cite{Girdhar2013IJRR} , a realtime online spatiotemporal topic modeling framework attempt to solve this problem of assigning high level labels to low level streaming observations.

Topic modeling techniques were originally developed for unsupervised semantic modeling of text documents \cite{Blei:2003} \cite{Griffiths:2004}. These algorithms automatically discover the main themes (topics) that underly these documents, which can then be used to compare these documents based on their semantic content.

\begin{figure}[t]
\begin{center}
\includegraphics[width=0.9\columnwidth]{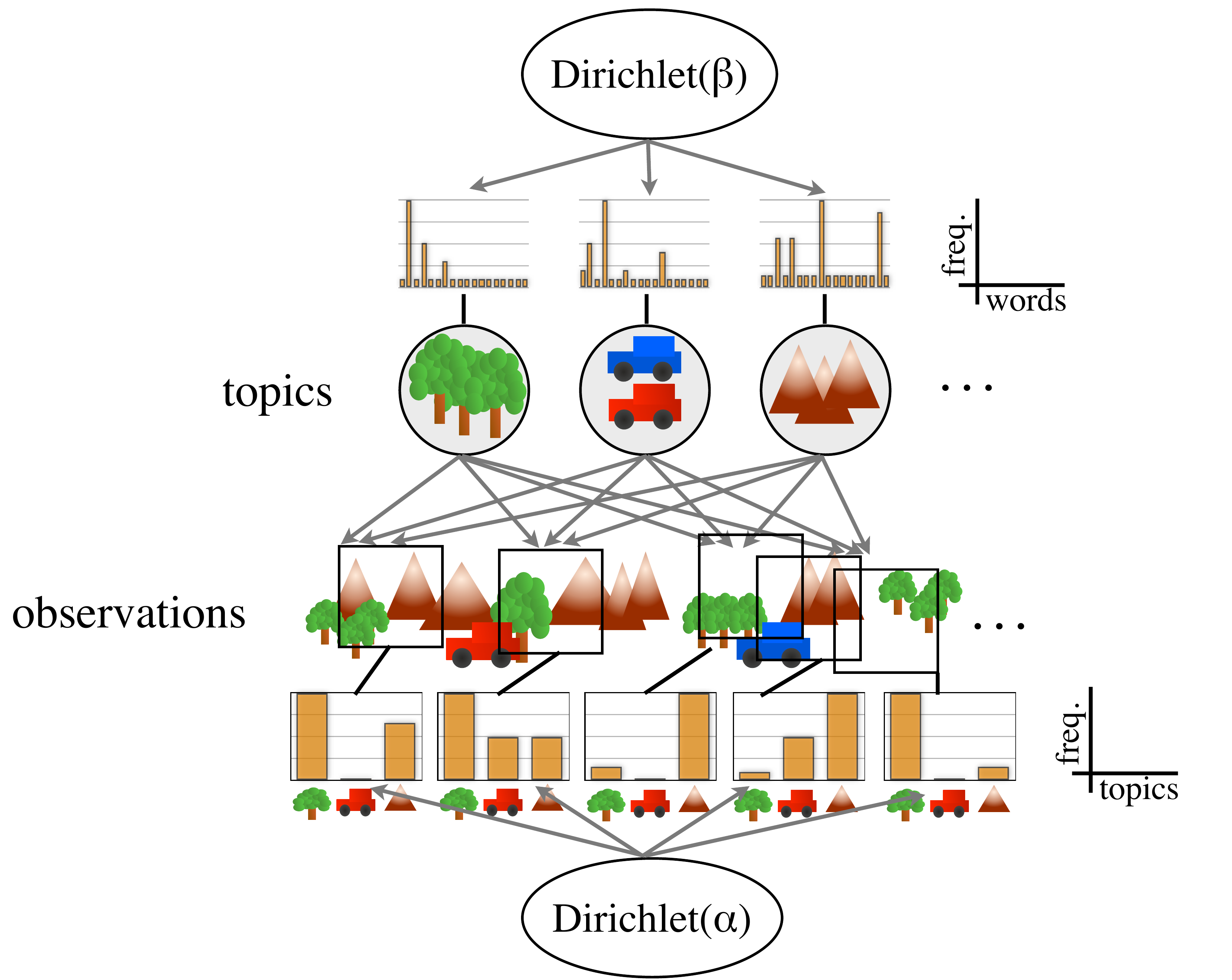}
\caption[Spatiotemporal topics]{Spatiotemporal Topics: As a robot observes the world, we would like its observations to be expressed as a mixture of topics with perceptual meaning. We model the topic distribution of all possible overlapping spatiotemporal regions or neighborhoods in the environment, and place a Dirichlet prior on their topic distribution. The topic distribution of the current observation can then be inferred given the topic labels for the neighborhoods in the view. Modeling neighborhoods allows us to use the context in which the current observation is being made to learn its topic labels. To guarantee realtime performance, we only refine a constant number of neighborhoods in each time step, giving higher priority to recently observed neighborhoods.}\label{fig:spatialtopics}

\end{center}
\end{figure}

Topic modeling of observation data captured by a mobile robot faces additional challenges compared to topic modeling of a collection of text documents, or images that are mutually independent.

\begin{itemize}
\item Robot observations are generally dependent on its location in space and time, and hence the corresponding semantic descriptor must take into account the location of the observed visual words during the refinement, and use it to compute topic priors that are sensitive to changes in time and the location of the robot.
\item The topic model must be updated online and in realtime, since the observations are generally made continuously at regular intervals. When computing topic labels for a new observation, we must also update topic labels for previous observations in the light on new incoming data.
\end{itemize}

ROST\cite{Girdhar2013IJRR} extends previous work on text and image topic modeling to make it suitable for processing streaming sensor data such as video and audio observed by a robot, and presents approximations for posterior inferencing that work in realtime. Topics in this case model the latent causes that produce these observations. ROST has been used for building semantic maps \cite{Girdhar2014ICRA} and for modeling curiosity in a mobile robot, for the purpose of information theoretic exploration \cite{Girdhar2014CRV}. ROST uses Gibbs sampling to continuously refine the topic labels for the observed data. In this paper we present various variants of Gibbs sampling that can be used to keep the topic labels converged under realtime constraints. 

\section{Previous Work}
	
\subsection{Topic Modeling of Spatiotemporal Data}
Given images of scenes with multiple objects, topic modeling has been used to discover objects in these images in an unsupervised manner. Bosch et al. \cite{Bosch:2006} used PLSA and a SIFT based~\cite{Lowe:IJCV:2004} visual vocabulary to model the content of images, and used a nearest neighbor classifier to classify the images. 

Fei-Fei et al.~\cite{FeiFei:2005:CVPR} have demonstrated the use of LDA to provide an intermediate representation of images, which was then used to learn an image classifier over multiple categories. 

Instead of modeling the entire image as a document, Spatial LDA (SLDA) \cite{Wang2007} models a subset of words, close to each other in an image as a document, resulting in a better encoding of the spatial structure. The assignment of words to documents is not done \emph{a priori}, but is instead modeled as an additional hidden variable in the generative process. 

Geometric LDA (gLDA) \cite{Philbin:2008:gLDA} models the LDA topics using words that are augmented with spatial position. Each topic in gLDA can be visualized as a pin-board where the visual words are pinned at their relatively correct positions. A document is assumed to be generated by first sampling a distribution over topics, and then for each word, sampling a topic label from this distribution, along with the transformation from the latent spatial model to the document (image). These transformations are all assumed to be affine, to model the change in viewpoints. 

LDA has been extended to learn a hierarchical representation of image content.  Sivic et al.\cite{Sivic:2008:CVPR} used hierarchical LDA (hLDA) \cite{Blei2004} for automatic generation of meaningful object hierarchies. Like LDA, hLDA also models documents as a mixture of topics; however, instead of the flat topics used in LDA, topics in hLDA correspond to a path in a tree. These topics become more specialized as they travel farther down from the root of the tree.

\section{Spatiotemporal Topic Model}
An observation word is a discrete observation made by a robot. Given the observation words and their location, we would like to compute the posterior distribution of topics at this location. Let $w$ be the observed word at location $x$. We assume the following probabilistic model for the observation words:

\begin{enumerate}
\item word distribution for each topic $k$: $$\phi_k \sim \mathrm{Dirichlet}(\beta),$$
\item topic distribution for words at location $x$ : $$\theta_{x} \sim \mathrm{Dirichlet}(\alpha + H(x)),$$
\item topic label for $w$: $$z \sim \mathrm{Discrete}(\theta_{x}),$$
\item word label: $$w \sim \mathrm{Discrete}(\phi_{z}),$$
\end{enumerate}
where $y\sim Y$ implies that random variable $y$ is sampled from distribution $Y$, $z$ is the topic label for the word observation $w$, and $H(x)$ is the distribution of topics in the neighborhood of location $x$. Each topic is modeled by distribution $\phi_k$ over $V$ possible word in the observation vocabulary. 

\begin{eqnarray}
\phi_k(v) = \Prob(w=v|z=k) = \propto  n^v_k + \beta,
\end{eqnarray}
where $n^v_k$ is the number of times we have observed word $v$ taking topic label $k$, and $\beta$ is the Dirichlet prior hyperparameter. Topic model $\Phi=\{\phi_k\}$ is a $K \times V$ matrix that encodes the global topic description information shared by all locations.

The main difference between this generative process and the generative process of words in a text document as proposed by LDA \cite{Blei:2003, Griffiths:2004} is in step 2. The context of words in LDA is modeled by the topic distribution of the document, which is independent of  other documents in the corpora. We relax this assumption and instead propose the context of an observation word to be defined by the topic distribution of its spatiotemporal neighborhood. This is achieved via the use of a kernel. The posterior topic distribution at location $x$ is thus defined as: 
\begin{eqnarray}
\theta_x(k) = \Prob(z=k|x) \propto \left( \sum_y K(x-y) n^k_y\right) + \alpha,
\end{eqnarray}
where $K(\cdot)$ is the kernel, $\alpha$ is the Dirichlet prior hyperameter and, $n^k_y$ is the number of times we observed topic $k$ at location $y$.

\section{Approximating Neighborhoods using Cells}
The generative process defined above models the clustering behavior of observations from a natural scene well, but is difficult to implement because it requires keeping track of the topic distribution at every location in the world. This is computationally infeasible for any large dataset. For the special case when the kernel is a uniform distribution over a finite region, we can assume a cell decomposition of the world, and approximate the topic distribution around a location by summing over topic distribution of cells in and around the location.

Let the world be decomposed into $C$ cells, in which each cell $c\in C$ is connected to its neighboring cells $G(c)~\subseteq~C$. Let $c(x)$ be the cell that contains points $x$. In this paper we only experiment with a grid decomposition of the world in which each cell is connected to its six nearest neighbors, 4 spatial and 2 temporal. However, the general ideas presented here are applicable to any other topological decomposition of spacetime. 


\begin{figure}[t]
\begin{center}
\includegraphics[width=0.6\columnwidth]{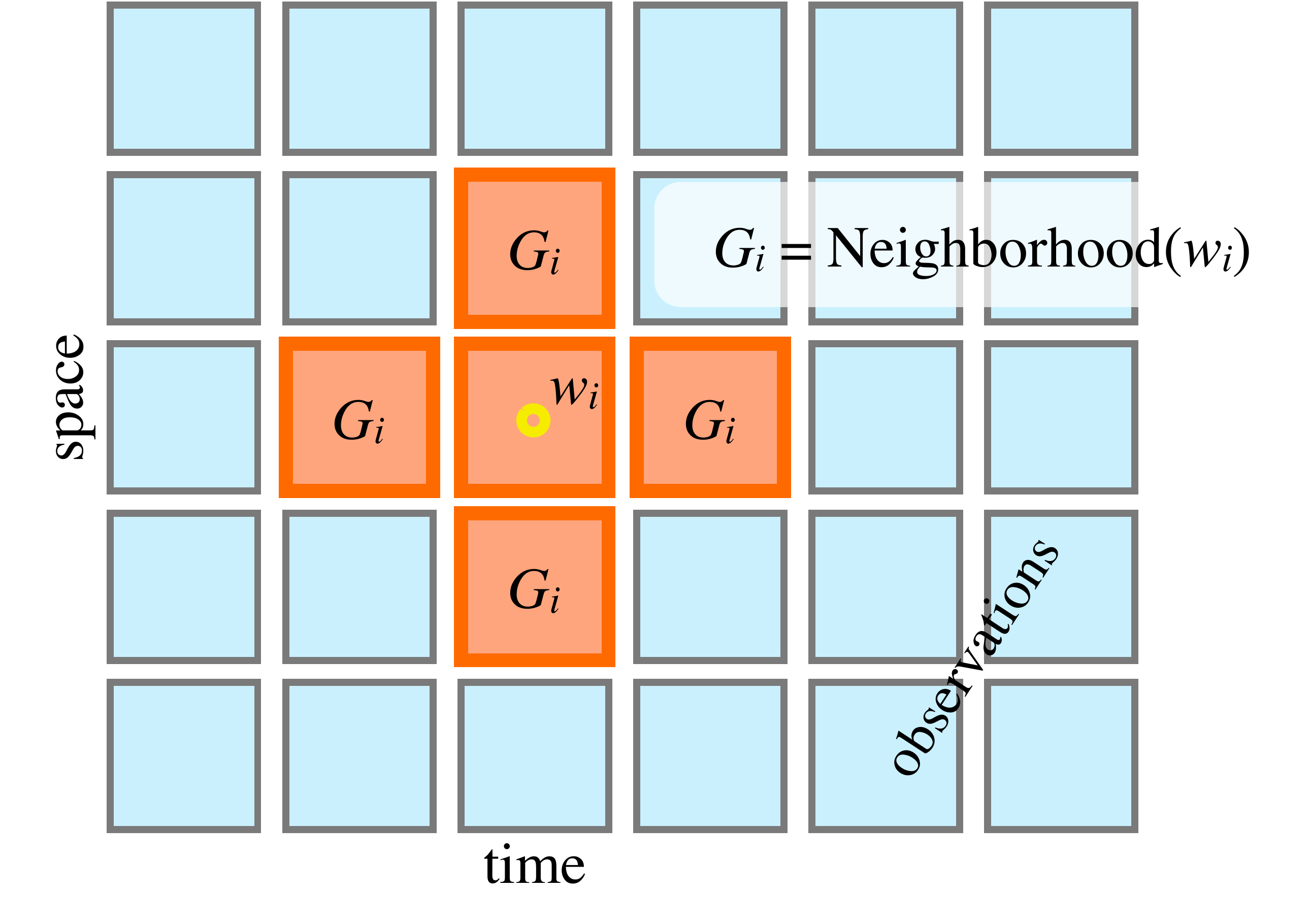}
\caption[Spatiotemporal neighborhood]{Each cell shown corresponds to a spatiotemporal bucket containing all the observation from that region. We refine the topic label for a word $w_i$ in an observation by taking into account the spatiotemporal context $G_i$ of the observation. }\label{fig:neighborhood}

\end{center}
\end{figure}

The topic distribution around $x$ can then be approximated using cells as: 
\begin{eqnarray}
\theta_x(k) \propto \left(\sum_{c' \in G(c(x))} n_{c'}^k\right) + \alpha
\end{eqnarray}

Due to this approximation, the following properties emerge:
\begin{enumerate}
\item $\theta_x = \theta_y$ if $c(x) = c(y)$, i.e., all the points in a cell share the same neighborhood topic distribution.
\item The topic distribution of the neighborhood is computed by summing over the topic distribution of the neighboring cells rather than individual points. 
\end{enumerate}
We take advantage of these properties while doing inference in realtime.

\section{Realtime Inference using Gibbs Sampling}

Given a word observation $w_i$, its location $x_i$, and its neighborhood $G_i = G(c(x_i))$, we use a Gibbs sampler to assign a new topic label to the word, by sampling from the posterior topic distribution:
\begin{eqnarray}
\begin{split}
\Prob(z_i = k |  w_i=v,x_i) \propto  \frac{n_{k,-i}^{v} + \beta}{\sum_{v=1}^V (n_{k,-i}^{v} + \beta)} \cdot \\
\frac{n_{G_i,-i}^{k} + \alpha}{\sum_{k=1}^K (n_{G_i,-i}^{k} + \alpha)},
\end{split} \label{eq:refine}
\end{eqnarray}
where $n_{k,-i}^{w}$ counts the number of words of type $w$ in topic $k$, excluding the current word $w_i$, $n_{G_i,-i}^{k}$ is the number of words with topic label $k$ in neighborhood $G_i$, excluding the current word $w_i$, and $\alpha, \beta$ are the Dirichlet hyper-parameters.  Note that for a neighborhood size of 0, the above Gibbs sampler is equivalent to the LDA Gibbs sampler proposed by Griffiths et al.\cite{Griffiths:2004}, where each cell corresponds to a document. Algorithm \ref{alg:batchgibbs} shows a simple iterative technique to compute the topic labels for the observed words in batch mode. 

\begin{algorithm}[t]	
	\DontPrintSemicolon
	\caption{Batch Gibbs sampling} 
	\label{alg:batchgibbs}
    Initialize $ \forall i, z_i \sim ~ \mathrm{Uniform}(\{1,\dots,K\})$ \;
    
	\While{true}{
  		\ForEach{ cell $c \in C$ }{   
	  		\ForEach{ word $w_i \in c $ }{   
	  			$z_i \sim \Prob(z_i = k |  w_i=v, x_i)$ \;
	  			Update $\Theta, \Phi$ given the new $z_i$ by updating $n_k^v$ and $n_G^k$\;
	  		}
        }	
	}
\end{algorithm}

In the context of robotics we are interested in the online refinement of observation data. After each new observation, we only have a constant amount of time to do topic label refinement. Hence, any online refinement algorithm that has computational complexity which increases with new data, is not useful. Moreover, if we are to use the topic labels of an incoming observation for making realtime decisions, then it is essential that the topic labels for the last observation converge before the next observation arrives. 

Since the total amount of data collected grows linearly with time, we must use a refinement strategy that efficiently handles global (previously observed) data and local (recently observed) data. 

Our general strategy is described by Algorithm \ref{alg:rostgibbs}. At each time step we add the new observations to the model, and then randomly pick observation times $t\sim\Prob(t|T)$, where $T$ is the current time, for which we resample the topic labels and update the topic model. 

\begin{algorithm}[t]	
	\DontPrintSemicolon
	\caption{Realtime Gibbs sampler} 
	\label{alg:rostgibbs} \label{alg:refinetopics}
	\While{true}{
	     Add new observed words to their corresponding cells.\;
	     $T \leftarrow 0$ (current time)\;
        Initialize $ \forall i\in M_T, z_i \sim ~ \mathrm{Uniform}(\{1,\dots,K\})$ \;
    
	    \While{no new observation}{
	        $t \sim \Prob(t|T)$\;
            \ForEach{ cell $c \in M_t$ }{   
			      \ForEach{ word $w_i \in c $ }{   
				    $z_i \sim \Prob(z_i = k |  w_i=v, x_i)$ \;
				    Update $\Theta, \Phi$ given the new $z_i$ by updating $n_k^v$ and $n_G^k$\;
			       }
            }	
        }
        $T \leftarrow T+1$\;	            
	}
\end{algorithm}

We discuss the choice of $\Prob(t|T)$ in the following sections.

\subsection{Now Gibbs Sampling}
The simplest way of processing streaming observation data to ensure that the topic labels from the last observation have converged is to only refine topics from the last observation till the next observation  has arrived. 
\begin{eqnarray}
\Prob(t|T) = 
\begin{cases}
1,& \text{if }t=T \\
0,& \text{otherwise} 
\end{cases}
\end{eqnarray}

We call this the Now Gibbs sampler. This is analogous to o-LDA approach by Banerjee and Basu \cite{Banerjee2007}.

If $R$ is our computation budget, defined as the expected number of observation time-steps our system can refine between the arrival times of two consecutive observations, and $r(t)$ be the number of times observations in $M_t$ have been refined after time $T$, then this approach gives each observation $R$ amount of resources.
\begin{eqnarray}
\Exp\{r(t)\} &=& R
\end{eqnarray}

Although this sounds fair, the problem is that no information from the future is used to improve the understanding of the past data.

\subsection{Uniform Gibbs Sampling}

A conceptually opposite strategy is to uniform randomly pick an observation from all the observations thus far, and refine the topic labels for all the words in this observation. 
\begin{eqnarray}
\Prob(t|T) = 1/T
\end{eqnarray}

This is analogous to the incremental Gibbs sampler for LDA proposed by Canini et al.\cite{Canini2009}.

Let $M_t$ be the set of cell containing observations at time $t$, $R$ be the number of observations our system can refine between two observations, and $r(t)$ be the number of times observations in $M_t$ have been refined after time $T$. The expected value of $r(t)$ is then:
\begin{eqnarray}
\Exp\{r(t)\} &=& R\left(\frac{1}{t} + \frac{1}{t+1} + \dots + \frac{1}{T}\right)\\
&\approx & R(\log T - \log t).
\end{eqnarray}

We see that older observations are sampled disproportionally higher than newer observations, and topic labels of new observations might take a long time to converge. In fact, if $\tau R$ is the expected number of iterations it takes for topic labels of an observation to converge, where $\tau < 1$ is a constant, then all observations after time $t' = 1/\tau$ would never be able to converge in the time before the next observation arrives. This is a big problem for a real-time system, where we need the topic labels of the last observations to actuate the robot. 

\subsection{Age Proportional Gibbs Sampling}

A seemingly good in-between approach might be to bias the random sampling of observations to be refined in favor of picking recent observations, with probability proportional to its timestamp. 
\begin{eqnarray}
\Prob(t|T) = \frac{t}{\sum_{i=1}^T i}
\end{eqnarray}

Then, the expected number of times this observation is refined is given by:
\begin{eqnarray}
\Exp\{r(t)\} &=& R \left(\frac{t}{\sum_{i=1}^t i} + \frac{t}{\sum_{i=1}^{t+1} i} + \dots +\frac{t}{\sum_{i=1}^T i }\right)\\
&\approx& 2R\frac{(T-t)}{T}.
\end{eqnarray}

When a new observation is made, the expected number of refinements it will gets before the next observation arrives is $R t/\sum t \approx 2R/t$, which implies that if $t'$ is the time after which it will not have sufficient number of refinements, then:
\begin{eqnarray}
\frac{2R}{t'} &=& \tau R \\
\implies t' &=& \frac{2}{\tau} 
\end{eqnarray}

Hence, we see that this strategy, although better than uniform random sampling (for which we computed $t' = 1/\tau$), is still not useful for long term operating of the robot.

\subsection{Exponential Gibbs Sampling}

Using a geometric distribution we can define the probability of refinement of timestep $t$, at current time $T$ 
\begin{eqnarray}
\Prob(t|T) = q(1-q)^{T-t},
\end{eqnarray}
where $0<q<1$ is a parameter. Using this distribution for picking refinement samples ensures that on average  $q R$  number of refinements are spent on refining the most recent observations, and the remaining $(q-1)R$ refinement iterations are spent on refining other recent observations. In the limit $T \rightarrow \infty$, observations in each time-step are refined $\Exp\{r(t)\}=R$ number of times, similar to Now Gibbs Sampler. This approach, however, allows new information to influence some of the recent past observations, resulting in lower global perplexity of the learned model.

\subsection{Mixed Gibbs Sampling}

We expect both Now and Exponential Gibbs samplers to be good at ensuring the topic labels for the last observation converges quickly (to a locally optimal solution), before the next observation arrives, whereas Uniform and Age-proportional Gibbs samplers are better at finding globally optimal results.

One way to balance both these performance goals is to combine these global and a local strategies. We consider four such approaches in this paper:

\begin{description}
\item[Uniform+Now:] 
\begin{eqnarray}
\Prob(t|T) = 
\begin{cases}
\eta,& \text{if }t=T \\
(1-\eta)/(T-1),& \text{otherwise} 
\end{cases}
\end{eqnarray}

\item[AgeProportional+Now:] 
\begin{eqnarray}
\Prob(t|T) = 
\begin{cases}
\eta,& \text{if }t=T \\
(1-\eta)\frac{t}{\sum_{i=1}^{T-1} i},& \text{otherwise} 
\end{cases}
\end{eqnarray}

\item[Uniform+Exp:]
\begin{eqnarray}
\Prob(t|T) = \eta q(1-q)^{T-t} + (1-\eta)/T
\end{eqnarray}

\item[AgeProportional+Exp:]
\begin{eqnarray}
\Prob(t|T) = \eta q(1-q)^{T-t} + (1-\eta)\frac{t}{\sum_{i=1}^{T} i}
\end{eqnarray}

\end{description}

Here $0 \leq \eta \leq 1$ is the mixing proportion between the local and the global strategies.

\section{Experiments}
\subsubsection{Dataset}
We evaluated the performance on ROST in analyzing videos using three different datasets with millions of visual words. We used a mixed vocabulary to describe each frame, with 5000 ORB words, 256 intensity words (pixel intensity), and 180 hue words (pixel hue), for a total vocabulary size of 5436. Although it is difficult to substantiate the optimality of the vocabulary, our experiments have suggested that once the vocabulary size is sufficiently large, there is limited sensitivity to its precise value~\cite{Girdhar:CRV:2011}. 

Some key statistics for these datasets is shown in Table \ref{tab:datasets_video_rost}. 

The \emph{2objects} dataset show a simple scenario in which two different objects appear on a textured (wood) background randomly, first individually and finally together.  

The \emph{aerial} dataset was collected using Unicorn UAV over a coastal region. The UAV performs a zig-zag coverage pattern over buildings, forested areas and ocean. 

The \emph{underwater} dataset was collected using Aqua as it swims over a coral reef. The dataset contains a variety of complex underwater terrain such as different coral species, rocks, sand, and divers. 

 The video files corresponding to these datasets, and some examples of ROST in action are available at \footnote{\url{http://cim.mcgill.ca/mrl/girdhar/}}.

\begin{table}[]
  \centering
  \begin{tabular}{|l|l|l|l|l|l|}
    \hline
    Name & size & T & N(words) & $\frac{\mathrm{N(words)}}{\mathrm{T}}$ & V\\
    \hline
    2objects &  720x480 & 1158 & 1741135 & 1503 & 5436\\
    aerial & 640x480 & 3600 & 8190231 & 2275 & 5436\\
    underwater & 1024x638 & 2569 & 4809869 & 1872 & 5436\\
    \hline
  \end{tabular}
  \caption[Video datasets for evaluating ROST]{Video datasets for evaluating ROST}
  \label{tab:datasets_video_rost}
\end{table}

To focus on analyzing the effects of spatiotemporal neighborhoods, and various Gibbs samplers, we fixed all other parameters of the system. We used cells of size 64x64 pixels with temporal width of 1 time step, Dirichlet parameters $\alpha=0.1, \beta=0.5$, number of topics $K=16$.

\subsection{Realtime Gibbs Samplers}
\begin{figure}
\centering
\subfigure[Refinement time $T_R=40$ ms.]{
\includegraphics[width=0.99\columnwidth]{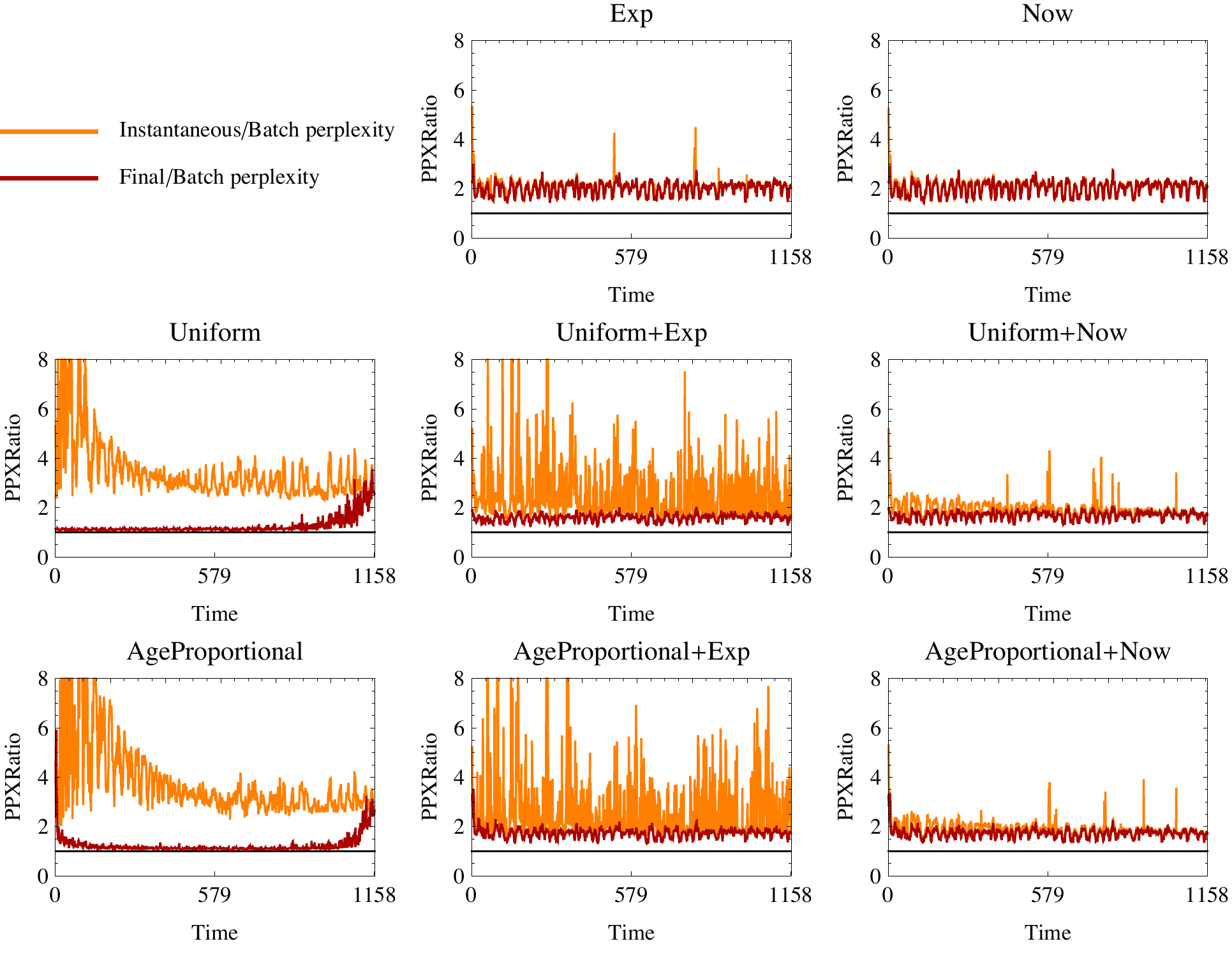}
}
\subfigure[Refinement time $T_R=160$ ms.]{
\includegraphics[width=0.99\columnwidth]{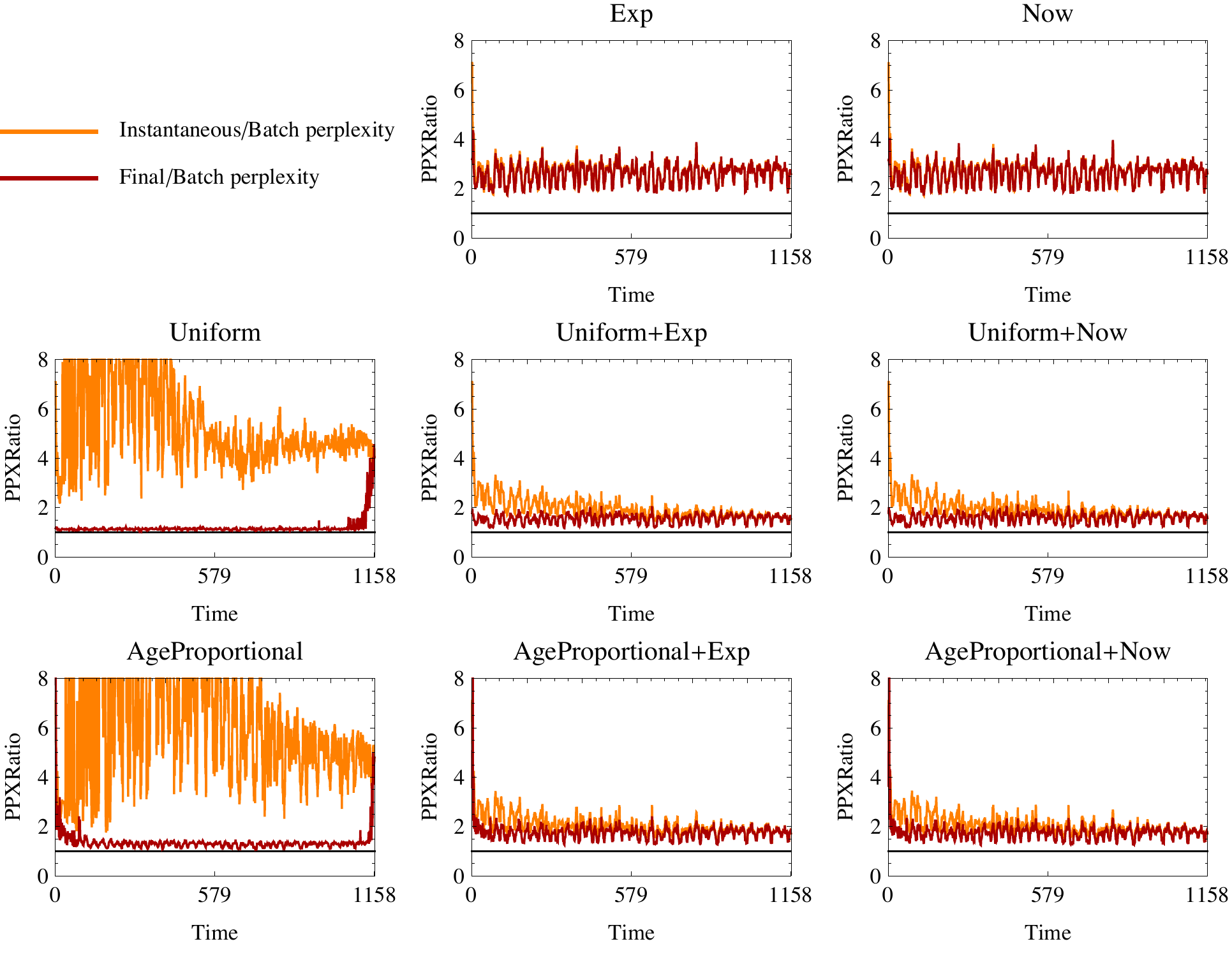}
}
\caption[Evaluating ROST Gibbs sampler - 2Objects dataset]{2Objects dataset --  ratio of instantaneous and final perplexity to batch perplexity, for each time step}\label{fig:2objects_gibbs}
\end{figure}

\begin{figure}
\centering
\subfigure[Refinement time $T_R=40$ ms.]{
\includegraphics[width=0.99\columnwidth]{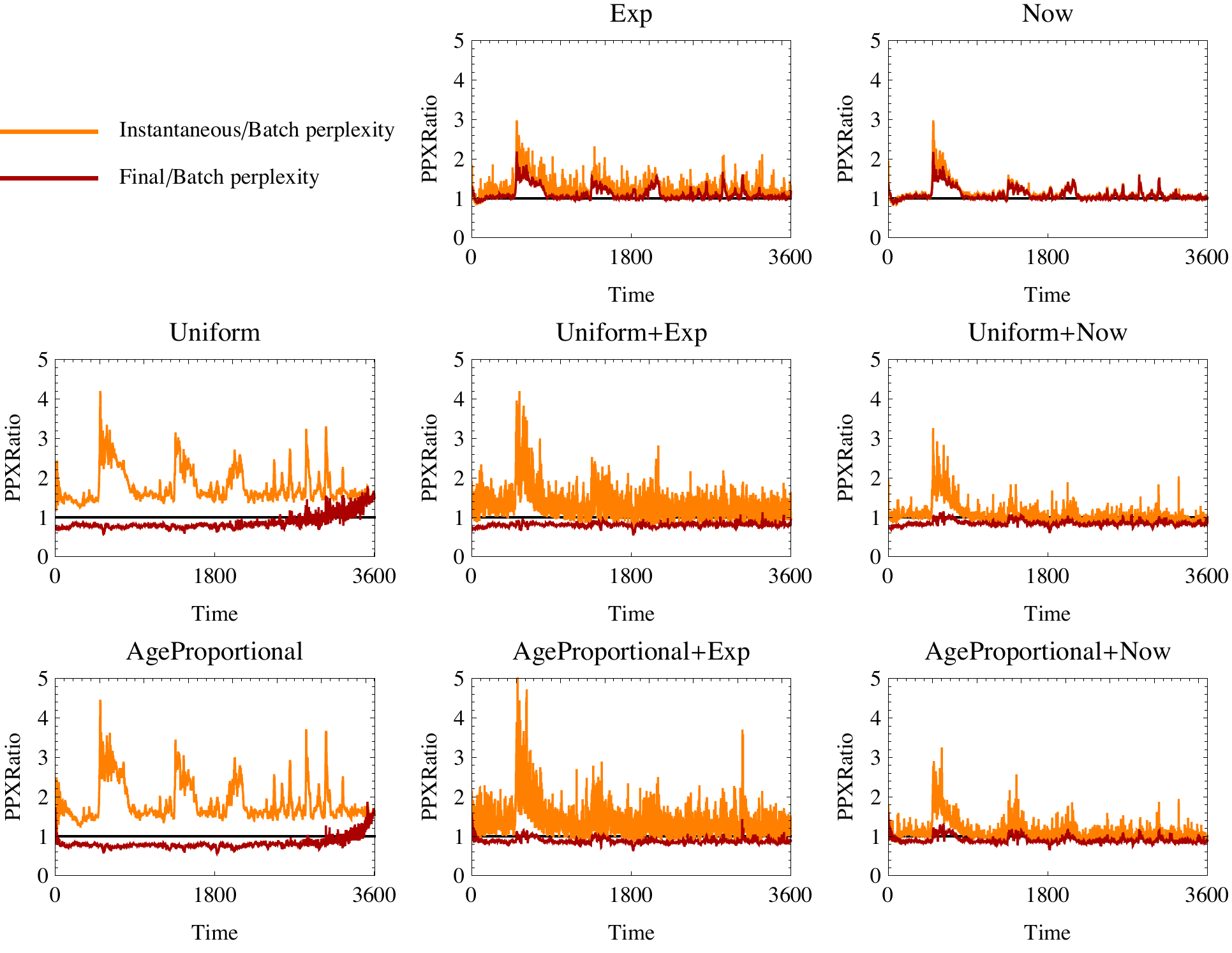}
}
\subfigure[Refinement time $T_R=160$ ms.]{
\includegraphics[width=0.99\columnwidth]{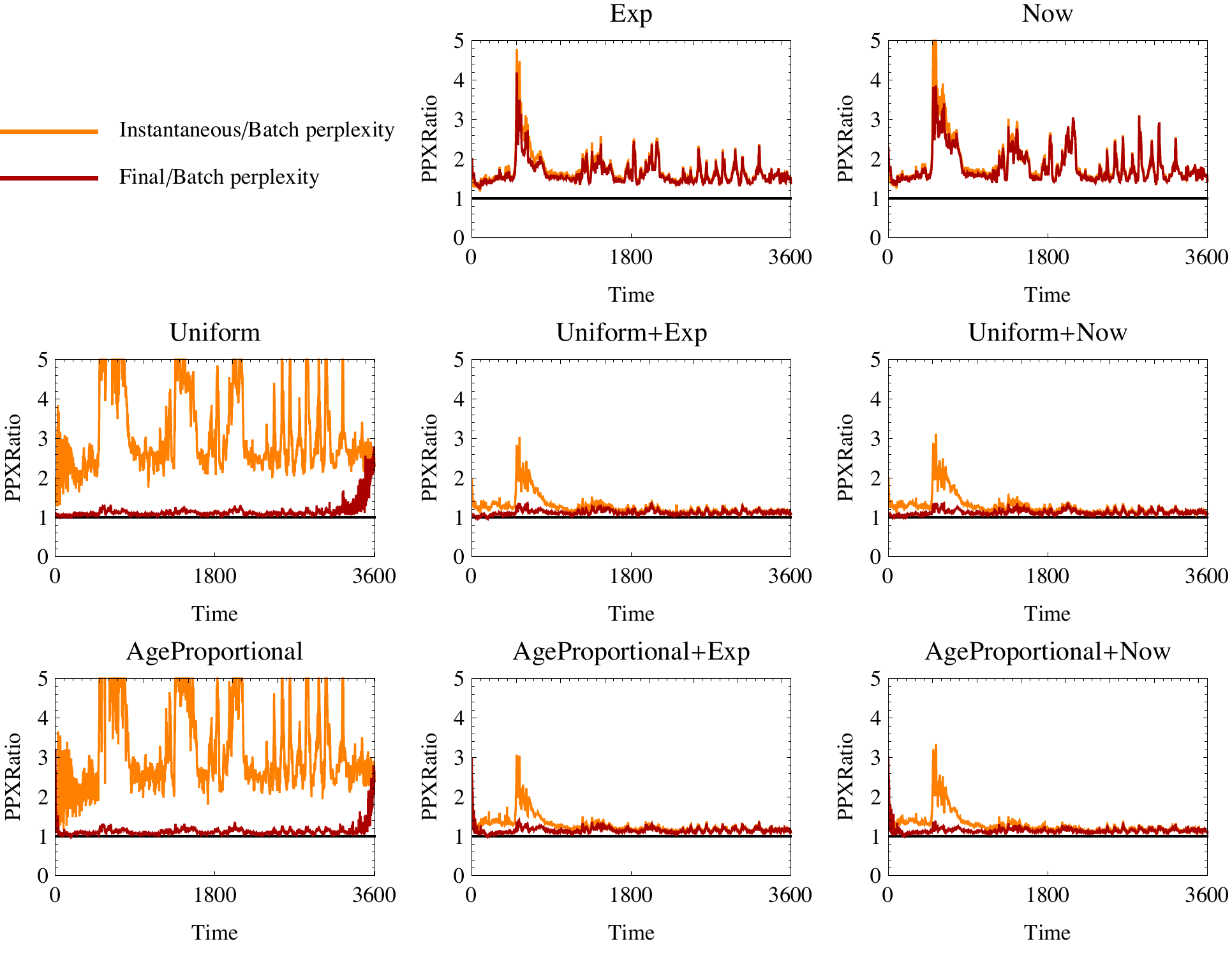}
}
\caption[Evaluating ROST Gibbs sampler - Aerial dataset]{Aerial dataset --  ratio of instantaneous and final perplexity to batch perplexity, for each time step}\label{fig:aerial_gibbs}
\end{figure}

\begin{figure}
\centering
\subfigure[Refinement time $T_R=40$ ms.]{
\includegraphics[width=0.99\columnwidth]{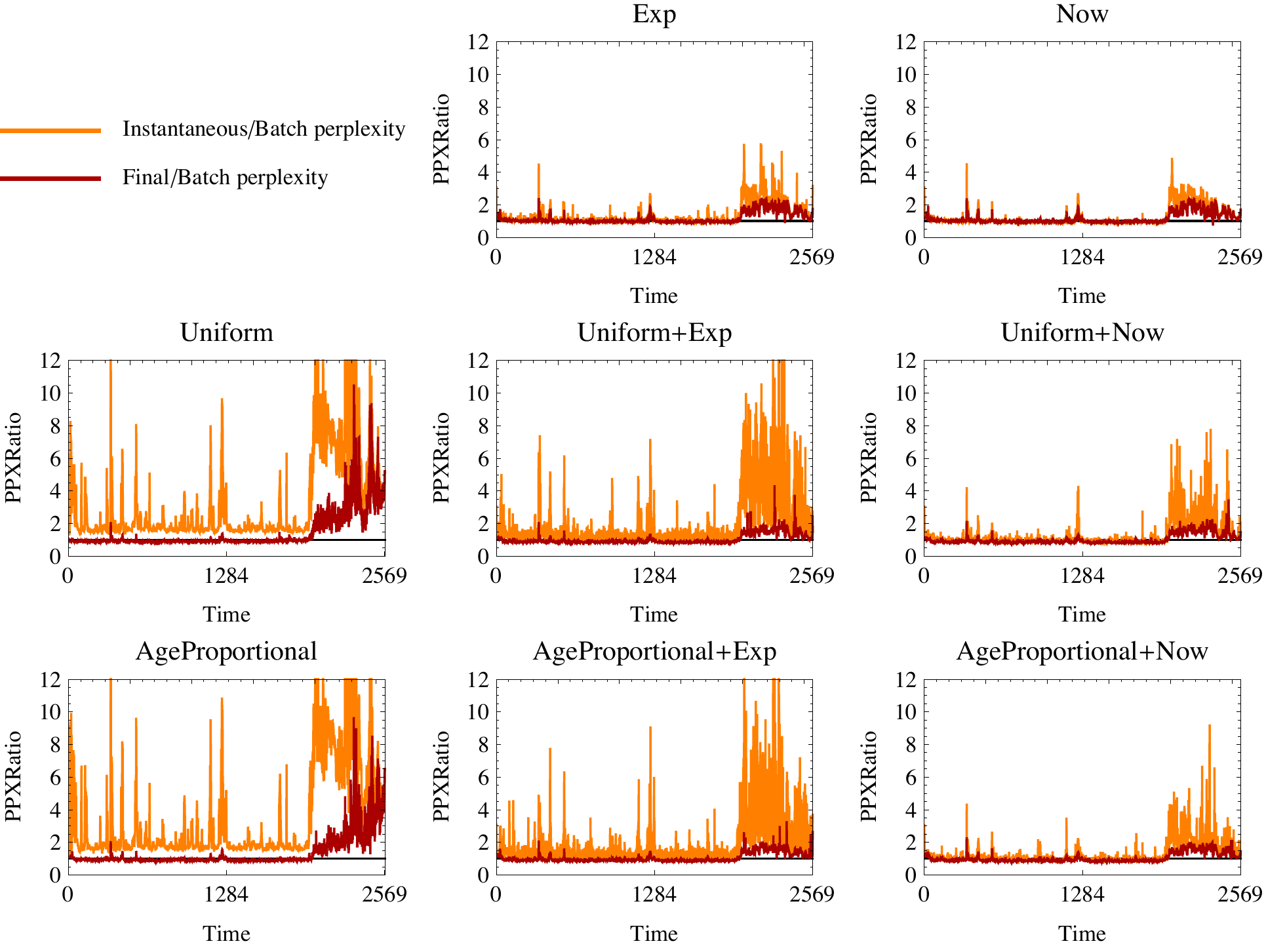}
}
\subfigure[Refinement time $T_R=160$ ms.]{
\includegraphics[width=0.99\columnwidth]{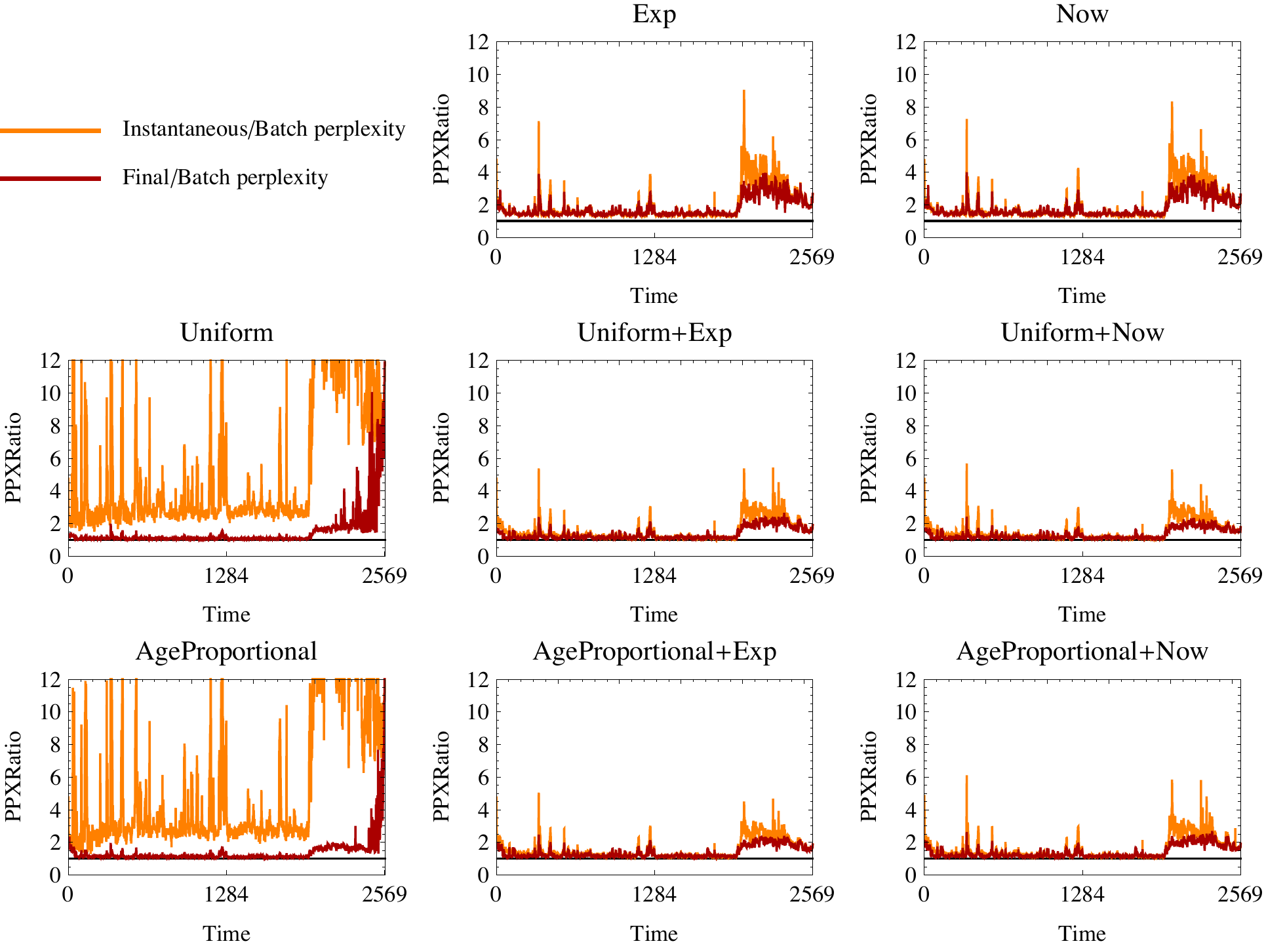}
}
\caption[Evaluating ROST Gibbs sampler - Underwater dataset]{Underwater dataset --  ratio of instantaneous and final perplexity to batch perplexity, for each time step}\label{fig:underwater_gibbs}
\end{figure}

To evaluate the proposed realtime Gibbs samplers on real data, we performed the following experiment. For each video dataset, and for each Gibbs sampler, we computed the topic labels and perplexity online, with 10 random restarts. We then compared the mean perplexity of words, one time step after their arrival (instantaneous), and after all observations have been made (final), with the perplexity of topic labels computed in batch. For a fair comparison, we used the same refinement time per time step ($T_R$) for both batch and online cases. The resulting perplexity plots are shown in Figures \ref{fig:2objects_gibbs}, \ref{fig:aerial_gibbs}, and \ref{fig:underwater_gibbs}. The mean perplexity scores for the entire datasets are shown in Tables \ref{tab:instantppx_video_rost} (instantaneous perplexity), and \ref{tab:finalppx_video_rost} (final perplexity). Note that instantaneous perplexity is computed on a new image, given the model learnt online from all previous data. Hence this perplexity score serves the same purpose as computing perplexity on held out data when evaluating topic modeling on batch data. 

\begin{table}[t]
  \centering
  \begin{tabular}{|l|c|c|c|c|c|c|}
    \hline
    Dataset $\rightarrow$ & \multicolumn{2}{c|}{2objects} & \multicolumn{2}{c|}{aerial} & \multicolumn{2}{c|}{underwater} \\
    \hline
    Alg.$\downarrow$, $T_R \rightarrow$ & 40 & 160 & 40 & 160 & 40 & 160 \\
    \hline
    Uniform  & 1.215 & \textbf{1.155} & 0.881 & 1.141 & 0.993 & \textbf{1.133} \\ 
    AgeP.    & \textbf{1.207} & 1.269 & 0.832 & 1.120 & 0.969 & 1.147 \\
    Exp.     & 1.799 & 2.272 & 1.065 & 1.548 & 1.005 & 1.488 \\
    Now      & 1.791 & 2.270 & 1.056 & 1.662 & 0.998 & 1.491 \\
    Uni+Now  & 1.555 & 1.420 & 0.842 & 1.109 & \textbf{0.885} & 1.152 \\
    AgeP+Now & 1.599 & 1.575 & 0.893 & 1.134 & 0.894 & 1.193\\
    Uni+Exp  & 1.480 & 1.398 & \textbf{0.808} & \textbf{1.1082} & 0.887 & 1.163 \\
    AgeP+Exp & 1.609 & 1.576 & 0.874 & 1.137 & 0.906 & 1.185 \\
    \hline
  \end{tabular}
  \caption{Mean final perplexity} 
  \label{tab:finalppx_video_rost}
\end{table}

\begin{table}[]
  \centering
  \begin{tabular}{|l|c|c|c|c|c|c|}
    \hline 
    Dataset $\rightarrow$ & \multicolumn{2}{c|}{2objects} & \multicolumn{2}{c|}{aerial} & \multicolumn{2}{c|}{underwater} \\
    \hline
    Alg.$\downarrow$, $T_R\rightarrow$ & 40 & 160 & 40 & 160 & 40 & 160 \\
    \hline
    Uniform &  3.500 & 4.826 & 1.665 & 2.798 & 1.907 & 3.140 \\ 
    AgeP. & 3.714 & 5.091 & 1.715 & 2.855 & 1.997 & 3.135 \\
    Exp. & 1.859 & 2.310 & 1.170 & 1.598 & 1.073 & 1.490 \\
    Now & 1.829  & 2.312 & 1.089 & 1.711 & 1.006 & 1.502 \\
    Uni+Now & 1.756 & \textbf{1.770} & \textbf{0.994} & 1.226 & \textbf{0.988} & \textbf{1.242} \\
    AgeP+Now & \textbf{1.739} & 1.883 & 1.028 & 1.247 & 0.996 & 1.270 \\
    Uni+Exp  & 2.267 & 1.784 & 1.279 & \textbf{1.217} & 1.357 & 1.249  \\
    AgeP+Exp & 2.320 & 1.873 & 1.339 & 1.238 & 1.375 & 1.256 \\
    \hline
  \end{tabular}
  \caption{Mean instantaneous perplexity perplexity}   \label{tab:instantppx_video_rost}
\end{table}

From our experiments we find that although Uniform and Age Proportional Gibbs samplers perform well when it comes to final perplexity of the dataset, they however perform poorly when measuring instantaneous perplexity. Low instantaneous perplexity, which is measured one time step after an observation is made, is essential for use of topic modeling in robotic applications. We would like to make  decisions based on current observations, and hence low instantaneous perplexity is crucial. We find that the mixed Gibbs samplers such as Uniform+Now perform consistently well. Note that all experiments with the mixed Gibbs samplers were performed with a fixed mixing ratio $\eta=0.5$, giving equal weight to local and global refinement. We are confident that better tuning of this variable will result in even better performance of ROST.

\section{Conclusion}
Topic modeling techniques such as ROST, model the latent context of the streaming spatiotemporal observation, such as image and other sensor data collected by a robot. 
In this paper we compared the performance of several Gibbs samplers for realtime spatiotemporal topic modeling, including those proposed by o-LDA and incremental LDA.

 We measured how well the topic labels converge, globally for the entire data, and for individually for an observation, one time step after its observation time. The latter measurement criterion is useful in evaluating the performance of the proposed technique in the context of robotics, where we need to make instantaneous decisions. We showed that the proposed mixed Gibbs samplers such as Uniform+Now perform consistently better than other samplers, which just focus on recent observation, or which refine all observation with equal probability.

\section*{Acknowledgment}
\small{This work was supported by the Natural Sciences and Engineering Research Council (NSERC) through the NSERC Canadian Field Robotics Network (NCFRN). Yogesh Girdhar is currently supported by the Postdoctoral Scholar Program at the Woods Hole Oceanographic Institution, with funding provided by the Devonshire Foundation and the J. Seward Johnson Fund. Authors would like to thank Julian Straub at MIT for helpful discussion. }

\bibliographystyle{IEEEtran}
\bibliography{IEEEabrv,library}
\end{document}